\documentclass[11pt]{article}

\usepackage[margin=1in]{geometry}
\usepackage[utf8]{inputenc}
\usepackage[T1]{fontenc}

\usepackage{times}
\usepackage{amsmath,amssymb}

\usepackage{graphicx}
\usepackage{booktabs}
\usepackage{multirow}
\usepackage{array}
\usepackage{xcolor}
\usepackage{colortbl}

\usepackage[colorlinks=true,linkcolor=blue!60!black,citecolor=green!50!black,urlcolor=blue!60!black]{hyperref}
\usepackage{natbib}

\usepackage{float}
\usepackage{placeins}
\usepackage{enumitem}
\usepackage{caption}
\usepackage{subcaption}

\newcommand{\ThermoQA}{\textsc{ThermoQA}}
\newcommand{\pp}{\,\text{pp}}
\newcommand{\cmark}{\ding{51}}
\newcommand{\xmark}{\ding{55}}

\usepackage{pifont}

\title{\ThermoQA{}: A Three-Tier Benchmark for Evaluating\\Thermodynamic Reasoning in Large Language Models}

\author{
  Kemal D\"uzkar\\
  Olivenet\\
  Kyrenia, Cyprus\\
  \texttt{kemal.duzkar@olivenet.io}
}

\date{March 2026}

\begin{document}

\maketitle

\begin{abstract}
We present \ThermoQA{}, a benchmark of 293 open-ended engineering thermodynamics problems in three tiers: property lookups (110~Q), component analysis (101~Q), and full cycle analysis (82~Q). Ground truth is computed programmatically from CoolProp~7.2.0, covering water, R-134a, and variable-$c_p$ air. Six frontier LLMs are evaluated across three independent runs each. The composite leaderboard is led by Claude Opus~4.6 (94.1\%), GPT-5.4 (93.1\%), and Gemini~3.1~Pro (92.5\%). Cross-tier degradation ranges from 2.8$\pp$ (Opus) to 32.5$\pp$ (MiniMax), confirming that property memorization does not imply thermodynamic reasoning. Supercritical water, R-134a refrigerant, and combined-cycle gas turbine analysis serve as natural discriminators with 40--60$\pp$ performance spreads. Multi-run $\sigma$ ranges from $\pm$0.1\% to $\pm$2.5\%, quantifying reasoning consistency as a distinct evaluation axis. Dataset and code are open-source at \url{https://huggingface.co/datasets/olivenet/thermoqa}.
\end{abstract}

\section{Introduction}
\label{sec:intro}

Engineering thermodynamics is a quantitative discipline where errors propagate through multi-step calculations. A 2\% error in enthalpy at the turbine inlet can cascade into a 10\% error in cycle thermal efficiency. This property makes thermodynamics an ideal testbed for evaluating LLM reasoning: unlike multiple-choice benchmarks where a model can guess or pattern-match, open-ended thermodynamic calculations require retrieval of accurate property data, correct application of conservation laws, and consistent multi-step algebraic reasoning.

Despite this, existing LLM benchmarks provide minimal coverage of engineering thermodynamics. General science benchmarks such as GPQA \citep{rein2023gpqa}, SciBench \citep{wang2023scibench}, and MMLU \citep{hendrycks2021mmlu} scatter thermodynamics across broader evaluation suites without isolating it as a coherent domain. Two dedicated thermodynamics benchmarks have recently appeared---UTQA \citep{geissler2025utqa} with 50 multiple-choice questions on ideal-gas processes, and \citet{loubet2025thermo} with 22 open-ended calculation problems on ideal fluids---but neither covers real-fluid properties (water, refrigerants), multi-component engineering systems, or exergy analysis. Crucially, no existing benchmark tests the full scope of an engineering thermodynamics course: from steam table lookups through component-level energy/entropy/exergy analysis to complete cycle calculations with real working fluids.

\ThermoQA{} fills this gap. Its 293 questions span property lookups, component analysis across seven device types, and full cycle calculations for ten cycle variants including Rankine, Brayton, vapor-compression refrigeration, and combined-cycle gas turbine systems. Four working fluids are tested (water, air, R-134a, and combined air+water), across three analysis depths (energy balance, entropy generation, and exergy destruction). Ground truth is computed programmatically from CoolProp and NASA polynomial correlations, ensuring reproducibility and eliminating human annotation error.

Our key contributions are:

\begin{enumerate}[leftmargin=*,topsep=2pt,itemsep=2pt]
\item \textbf{Tiered benchmark design.} Three tiers test distinct capabilities: property memorization (Tier~1), multi-step component reasoning (Tier~2), and full-system cycle analysis (Tier~3). Rankings reshuffle across tiers---Gemini leads Tier~1 but falls to \#3 on Tier~3, while Opus climbs from \#3 to \#1---confirming that these are separable skills.

\item \textbf{Programmatic ground truth.} All 293 answers are computed from CoolProp~7.2.0 (IAPWS-IF97 for water, Helmholtz EOS for R-134a) and NASA 7-coefficient polynomials for air, cross-verified to $<$0.01\% against NIST reference data. No hand-authored solutions.

\item \textbf{Multi-run consistency analysis.} Three independent runs per model reveal standard deviations ranging from $\pm$0.1\% (GPT-5.4 on Tier~3) to $\pm$2.5\% (DeepSeek-R1 on Tier~2), quantifying reasoning reliability as a distinct evaluation dimension.

\item \textbf{Natural discriminators.} We identify supercritical water, R-134a, compressor analysis, variable-$c_p$ Brayton cycles, and CCGT as question subsets that produce the widest performance spreads, offering diagnostic value for model developers.

\item \textbf{Open-source release.} Dataset, code, evaluation scripts, and all model responses are released under CC-BY-4.0 / MIT at \url{https://huggingface.co/datasets/olivenet/thermoqa} and \url{https://github.com/olivenet-iot/ThermoQA}.
\end{enumerate}

\section{Related Work}
\label{sec:related}

\paragraph{General science benchmarks.} GPQA \citep{rein2023gpqa} offers 448 graduate-level MCQs in biology, physics, and chemistry, but does not isolate engineering thermodynamics as a coherent domain. \citet{geissler2025utqa} note that GPQA's chemistry coverage is dominated by organic chemistry, with almost no treatment of entropy or reversibility. SciBench \citep{wang2023scibench} provides 789 collegiate-level open-ended problems; its thermodynamics items are limited to end-answer calculations without probing multi-step reasoning about state functions or reversibility. MMLU \citep{hendrycks2021mmlu} treats thermodynamics as scattered MCQ items across its 57 subjects. None isolate engineering thermodynamics as a coherent evaluation domain.

\paragraph{Engineering-specific benchmarks.} EngiBench \citep{zhou2025engibench} evaluates LLMs on hierarchical engineering problems across three difficulty levels, finding sharp performance drops on open-ended tasks and significant sensitivity to numerical perturbations. MatSciBench \citep{matscibench2025} targets materials science with 1,340 problems. These cover engineering broadly but not thermodynamic cycle analysis, real-fluid properties, or exergy.

\paragraph{Thermodynamics-specific work.} \textbf{UTQA} \citep{geissler2025utqa} is a 50-item single-choice benchmark (33 text-only, 17 diagram-based) targeting ideal-gas processes, reversibility, and diagram interpretation. Across 19 models, the best overall score is 82\% (gpt-o3); no model reaches the authors' 95\% tutoring-reliability threshold. Two failure modes emerge: fragile regime recognition on finite-rate/irreversible scenarios, and a multimodal binding deficit on diagrams where mean accuracy drops to 32\%. The authors explicitly scope their benchmark as ``intentionally narrow (ideal gases; excluding real-gas effects, mixtures, phase equilibria)'' and identify extensions toward real-gas behavior and standard cycles as future work---precisely the domain \ThermoQA{} addresses.

\textbf{\citet{loubet2025thermo}} evaluate five LLMs on 22 calculation problems with expert human grading, restricted to ideal fluids (ideal gases and constant-density liquids). They find substantial run-to-run variability (standard deviations of 1/3 to 1/2 of mean score), which we corroborate. In a follow-up, \citet{loubet2025superstudent} report that o3 outperformed all 90 students on a university thermodynamics exam in zero-shot mode, achieving a score in the range of the best results observed across more than 10{,}000 similar exams since 1985. \citet{shahid2026bloom} apply Bloom's taxonomy to evaluate ChatGPT across chemical engineering education, finding a clear performance gradient from $\sim$95\% on factual recall to $\sim$41\% on creative/evaluative tasks.

\paragraph{Positioning.} Table~\ref{tab:comparison} summarizes the landscape. \ThermoQA{} differs from all prior work in three dimensions: \emph{format} (open-ended numerical calculation with step-level weighted scoring, eliminating the MCQ guessing advantage), \emph{fluid coverage} (water via IAPWS-IF97, R-134a via Helmholtz EOS, and variable-$c_p$ air via NASA polynomials---the real-fluid extensions that UTQA identifies as future work), and \emph{scope} (property lookups through full CCGT cycle analysis with exergy destruction). Ground truth is programmatic via CoolProp, not human-authored, enabling indefinite generation of new problem instances.

\begin{table}[t]
\centering
\caption{Comparison with existing thermodynamics-related LLM benchmarks. \ThermoQA{} is the first to combine open-ended numerical calculation with real-fluid coverage, full cycle analysis, and exergy.}
\label{tab:comparison}
\small
\begin{tabular}{@{}lccccccc@{}}
\toprule
\textbf{Benchmark} & \textbf{Size} & \textbf{Format} & \textbf{Scoring} & \textbf{Real Fluids} & \textbf{Cycles}$^\ddagger$ & \textbf{Exergy} & \textbf{Multi-run} \\
\midrule
GPQA \citeyearpar{rein2023gpqa} & 448$^\dagger$ & MCQ & Exact match & \xmark & \xmark & \xmark & \xmark \\
SciBench \citeyearpar{wang2023scibench} & 789$^\dagger$ & Open & Final answer & \xmark & \xmark & \xmark & \xmark \\
UTQA \citeyearpar{geissler2025utqa} & 50 & MCQ & Exact match & \xmark & \xmark & \xmark & \xmark \\
Loubet et al.\ \citeyearpar{loubet2025thermo} & 22 & Open calc. & Expert grading & \xmark & \cmark$^*$ & \xmark & \cmark \\
EngiBench \citeyearpar{zhou2025engibench} & 1293$^\dagger$ & Mixed & Binary/rubric & \xmark & \xmark & \xmark & \xmark \\
\textbf{\ThermoQA{} (ours)} & \textbf{293} & \textbf{Open calc.} & \textbf{Step-weighted} & \textbf{\cmark} & \textbf{\cmark} & \textbf{\cmark} & \textbf{\cmark} \\
\bottomrule
\end{tabular}
\vspace{2pt}

{\footnotesize $^\dagger$Total benchmark size; thermodynamics subset is smaller. $^*$Ideal gas cycles only. $^\ddagger$Full numerical cycle calculations (Rankine, Brayton, VCR, CCGT); conceptual cycle questions are not counted.}
\end{table}

\section{Benchmark Design}
\label{sec:design}

\subsection{Tier Structure}

\ThermoQA{} comprises three tiers of increasing thermodynamic complexity. Each tier builds on the skills required by the previous one.

\textbf{Tier 1: Property Lookups (110 questions).} Given thermodynamic state specifications (e.g., ``water at 5 MPa and 400$^\circ$C''), models must report thermodynamic properties---specific enthalpy $h$, specific entropy $s$, specific volume $v$, internal energy $u$, density $\rho$, and/or quality $x$---and identify the phase region. Eight categories span the phase diagram: subcooled liquid (10~Q), saturated liquid (12~Q), wet steam (18~Q), saturated vapor (10~Q), superheated vapor (20~Q), supercritical (10~Q), phase determination (15~Q), and inverse lookups (15~Q). Difficulty is assigned as easy/medium/hard based on proximity to phase boundaries and the critical point.

\textbf{Tier 2: Component Analysis (101 questions).} Each question specifies a thermodynamic component (turbine, compressor, pump, heat exchanger, boiler, mixing chamber, or nozzle) with inlet/outlet conditions and isentropic efficiency. Models perform multi-step analysis at three depths: (A) energy balance only, (B) energy + entropy generation, (C) energy + entropy + exergy destruction. Three fluids: water (74~Q), air (17~Q), R-134a (10~Q). Each answer includes 5--12 numerical steps, scored with step-level weights reflecting engineering importance.

\textbf{Tier 3: Cycle Analysis (82 questions).} Full thermodynamic cycle calculations across 10 cycle types: ideal/actual/reheat Rankine (RNK-I/A/RH), ideal/actual/regenerative Brayton (BRY-I/A/RG), aftercooling variable-$c_p$ Brayton (BRY-AV), regenerative variable-$c_p$ Brayton (BRY-RV), actual vapor-compression refrigeration (VCR-A), and combined-cycle gas turbine (CCGT). Four fluid contexts: water (27~Q), air (28~Q), R-134a (15~Q), and combined air+water (12~Q). Each answer includes 15--60 numerical steps with 6-tier weighting (1--6).

\subsection{Ground Truth Generation}

All reference values are computed programmatically:

\begin{itemize}[leftmargin=*,topsep=2pt,itemsep=1pt]
\item \textbf{Water/steam:} CoolProp 7.2.0 \citep{coolprop} with IAPWS-IF97 backend \citep{iapws_if97}, validated against NIST to $<$0.037\% maximum deviation across 13 saturation points.
\item \textbf{R-134a:} CoolProp Helmholtz EOS with IIR reference state ($h=200$~kJ/kg, $s=1.0$~kJ/(kg$\cdot$K) at 0$^\circ$C saturated liquid), specified explicitly in each question to avoid convention mismatch.
\item \textbf{Air:} NASA 7-coefficient polynomials matching \citet{cengel2019} Table~A-17 for variable specific heats. This is critical: CoolProp's ``Air'' models a real gas mixture (N$_2$/O$_2$/Ar), producing $\sim$126~kJ/kg offsets versus textbook ideal-gas values. Using the NASA polynomial ensures consistency with standard engineering education.
\item \textbf{Dead state:} $T_0 = 25^\circ$C (298.15~K), $P_0 = 0.1$~MPa for all exergy calculations.
\end{itemize}

Question parameters (pressures, temperatures, efficiencies, mass flow rates) are sampled from physically realistic ranges and validated against CoolProp to ensure all states are thermodynamically feasible. An anchor-derive pattern generates inlet states, then derives outlet states via conservation equations---identical to the procedure a student would follow.

\subsection{Scoring}

\textbf{Tier 1.} Per-property scoring: $\pm$2\% relative tolerance OR $\pm$0.5 absolute (whichever more lenient). The 2\% threshold matches standard engineering homework tolerance \citep{cengel2019} and exceeds the $<$0.1\% uncertainty of IAPWS-IF97 itself, ensuring we measure model error rather than reference uncertainty. Quality $x$: $\pm$0.03 absolute. Phase: exact string match against an alias list.

\textbf{Tiers 2--3.} Weighted step-level scoring:
\begin{equation}
\text{Score}_q = \frac{\sum_{i=1}^{N} w_i \cdot \mathbb{1}[\text{step}_i \text{ correct}]}{\sum_{i=1}^{N} w_i}
\end{equation}
where $w_i$ is the weight of step $i$ and correctness uses the same $\pm$2\%/$\pm$0.5 tolerance. Weights reflect engineering importance: inlet properties carry low weights ($w = 1$--$2$), intermediate calculations carry medium weights ($w = 2$--$4$), and final engineering outcomes ($\dot{W}_\text{net}$, $\eta_\text{th}$, $\dot{X}_\text{dest}$, $\eta_\text{II}$) carry the highest weights ($w = 4$--$6$). This design ensures that a model which correctly retrieves inlet properties but fails the actual engineering analysis receives a low score. Dimensionless quantities ($\eta_\text{th}$, $\eta_\text{II}$, COP) use a tighter absolute tolerance of $\pm$0.02, since the default $\pm$0.5 absolute tolerance was found to pass values with $>$20\% error on quantities in the $[0,1]$ range.

\subsection{Evaluation Pipeline}

The evaluation pipeline has four stages:

\begin{enumerate}[leftmargin=*,topsep=2pt,itemsep=1pt]
\item \textbf{Question delivery.} Each question is submitted via API with a system prompt requesting ``symbol = value unit'' format. No few-shot examples are provided---this is a zero-shot evaluation, testing the model's intrinsic thermodynamic knowledge rather than its ability to learn from examples.

\item \textbf{Model response.} Models respond in free-form text with all reasoning/thinking modes enabled at their highest settings. No constraints are placed on response format.

\item \textbf{Two-pass extraction.} First, regex patterns extract numerical values from the response. Second, gpt-4.1-mini (temperature=0) re-extracts answers from the full response text, including any thinking/reasoning tokens exposed by the API. A \emph{take-max} strategy combines both passes: for each scored step, we retain whichever extraction yields a correct answer. This ensures the extraction pipeline never degrades performance relative to regex alone---we test thermodynamic knowledge, not output formatting. The choice of gpt-4.1-mini as extractor reflects its low cost, fast inference, and near-deterministic behavior at temperature=0.

\item \textbf{Scoring.} Extracted values are compared against CoolProp ground truth using the tolerance scheme described above.
\end{enumerate}

\section{Experimental Setup}
\label{sec:setup}

\subsection{Models}

We evaluate six frontier LLMs spanning five providers:

\begin{table}[h]
\centering
\small
\begin{tabular}{@{}llll@{}}
\toprule
\textbf{Model} & \textbf{Provider} & \textbf{API Model String} & \textbf{Reasoning} \\
\midrule
Claude Opus 4.6 & Anthropic & claude-opus-4-6 & Adaptive \\
GPT-5.4 & OpenAI & gpt-5.4 & reasoning\_effort=high \\
Gemini 3.1 Pro & Google & gemini-3.1-pro-preview & thinking\_level=HIGH \\
Grok 4 & xAI & grok-4.20-beta-0309 & Native \\
DeepSeek-R1 & DeepSeek & deepseek-reasoner & Native \\
MiniMax M2.5 & MiniMax & MiniMax-M2.5 & Inline \\
\bottomrule
\end{tabular}
\end{table}

\subsection{Prompt Design}

All models receive an identical system prompt instructing them to show their reasoning and report final answers in ``symbol = value unit'' format. Each question is submitted as a single-shot query with no few-shot examples. Temperature is set to the provider default for reasoning models (typically 1.0); all reasoning/thinking modes are enabled at their highest setting. The full system prompt and an example API call are provided in the supplementary repository.

We use a single canonical prompt rather than testing multiple prompting strategies. \citet{geissler2025utqa} evaluated 17 prompting strategies on their UTQA benchmark and found that while prompt choice produced real accuracy differences, run-to-run standard deviation was only $\sigma \approx 0.05$, and no consistent ordering of strategies emerged---suggesting that prompt design primarily affects surface presentation rather than correcting deeper deficits in scientific reasoning. We therefore prioritize breadth of models and tiers over prompt-variant analysis, and leave prompt sensitivity studies to future work.

\subsection{Multi-Run Protocol}

Each model is evaluated three times independently per tier ($3 \times 6 \times 3 = 54$ evaluation runs, $\sim$5{,}300 individual API calls). Results are reported as mean $\pm$ standard deviation. Three runs is the minimum for computing standard deviation and sufficient to separate most model pairs via Welch's $t$-test: 13 of 15 pairwise comparisons on Tier~3 reach significance at $\alpha = 0.05$. This matches the protocol of \citet{loubet2025thermo}, who also use three independent runs per model. The composite score weights tiers by question count:
\begin{equation}
\text{Composite} = \frac{110 \times \bar{S}_1 + 101 \times \bar{S}_2 + 82 \times \bar{S}_3}{293}
\end{equation}


\FloatBarrier
\section{Results}
\label{sec:results}

\subsection{Overall Leaderboard}

\begin{table}[t]
\centering
\caption{Final leaderboard (3-run mean $\pm$ std). Composite = question-weighted average.}
\label{tab:leaderboard}
\small
\begin{tabular}{@{}clcccccc@{}}
\toprule
\textbf{Rank} & \textbf{Model} & \textbf{Tier 1} & \textbf{Tier 2} & \textbf{Tier 3} & \textbf{T1$\to$T3 Drop} & \textbf{Composite} \\
\midrule
1 & Claude Opus 4.6 & 96.4 $\pm$ 0.9 & \textbf{92.1 $\pm$ 0.1} & \textbf{93.6 $\pm$ 0.5} & $-$2.8$\pp$ & \textbf{94.1\%} \\
2 & GPT-5.4 & 97.8 $\pm$ 0.8 & 90.8 $\pm$ 0.5 & 89.7 $\pm$ 0.1 & $-$8.1$\pp$ & 93.1\% \\
3 & Gemini 3.1 Pro & \textbf{97.9 $\pm$ 0.5} & 90.8 $\pm$ 1.2 & 87.5 $\pm$ 1.5 & $-$10.4$\pp$ & 92.5\% \\
4 & DeepSeek-R1 & 90.5 $\pm$ 0.2 & 89.2 $\pm$ 2.5 & 81.0 $\pm$ 2.2 & $-$9.5$\pp$ & 87.4\% \\
5 & Grok 4 & 91.8 $\pm$ 1.2 & 87.9 $\pm$ 0.7 & 80.4 $\pm$ 0.8 & $-$11.4$\pp$ & 87.3\% \\
6 & MiniMax M2.5 & 85.2 $\pm$ 0.6 & 76.2 $\pm$ 1.1 & 52.7 $\pm$ 1.5 & $-$32.5$\pp$ & 73.0\% \\
\bottomrule
\end{tabular}
\end{table}

\begin{figure}[!htb]
\centering
\includegraphics[width=0.95\linewidth]{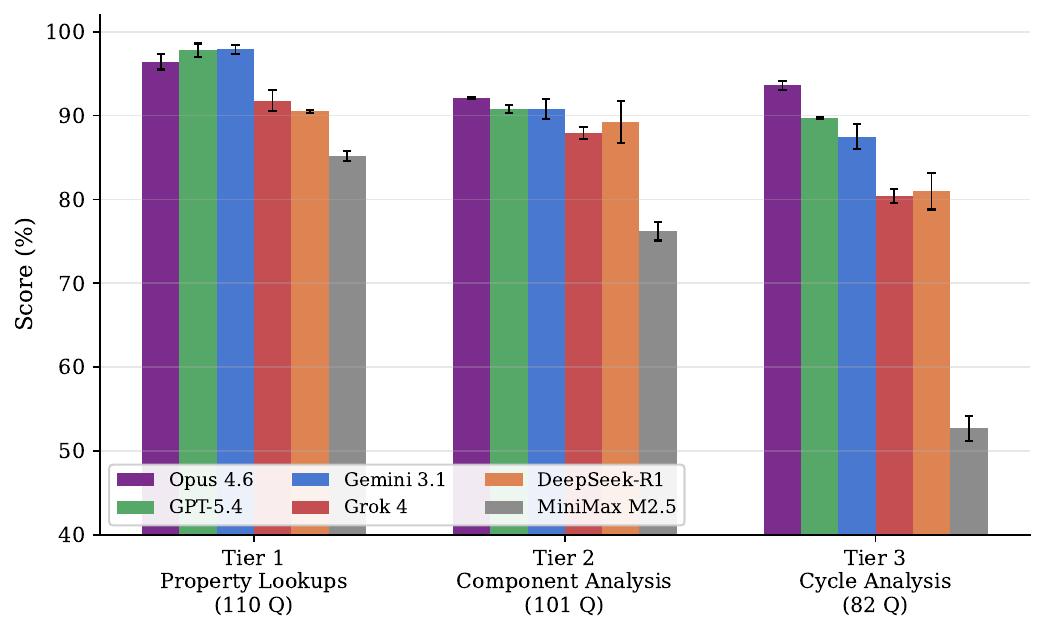}
\caption{Per-tier performance for all six models (3-run mean with $\pm\sigma$ error bars). Tier difficulty increases left to right. Note the ranking reshuffle: Gemini leads Tier~1 but Opus leads Tiers~2--3.}
\label{fig:cross_tier}
\end{figure}

Table~\ref{tab:leaderboard} presents the final rankings. The performance spread is 21.1 percentage points between the top (94.1\%) and bottom (73.0\%) models, confirming that engineering thermodynamics is a meaningful differentiator. Figure~\ref{fig:cross_tier} visualizes the per-tier scores.

\subsection{Cross-Tier Degradation}

\begin{figure}[!htb]
\centering
\includegraphics[width=0.85\linewidth]{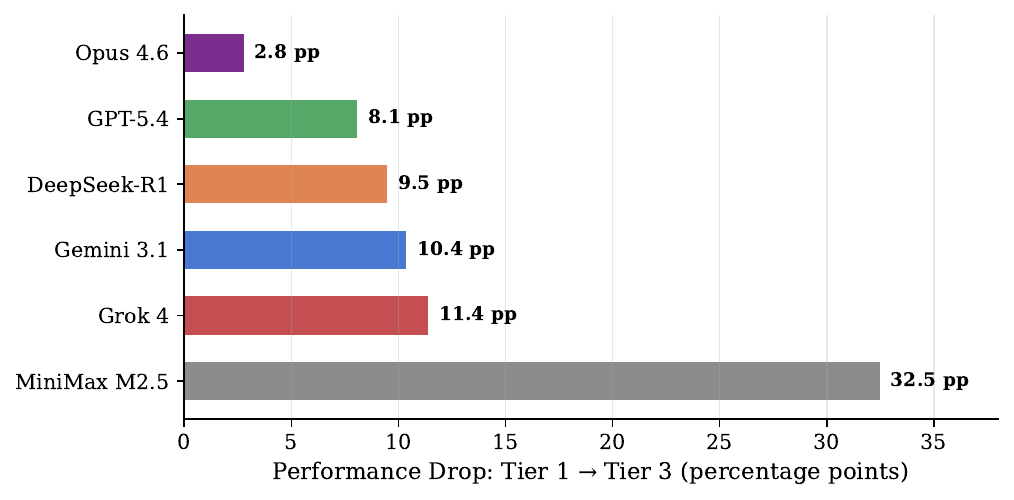}
\caption{Cross-tier degradation: performance drop from Tier~1 (property lookups) to Tier~3 (cycle analysis). Smaller is better.}
\label{fig:degradation}
\end{figure}

The performance drop from Tier~1 to Tier~3 quantifies how well property-lookup skill translates into reasoning ability. Figure~\ref{fig:degradation} shows that Opus degrades only 2.8$\pp$ (96.4\%$\to$93.6\%), while MiniMax drops 32.5$\pp$ (85.2\%$\to$52.7\%). This metric is arguably more informative than absolute accuracy: a model that excels at memorization but fails at reasoning is less useful for engineering applications than one with consistent performance across complexity levels.

\subsection{Discriminator Analysis}

\begin{figure}[!htb]
\centering
\includegraphics[width=0.95\linewidth]{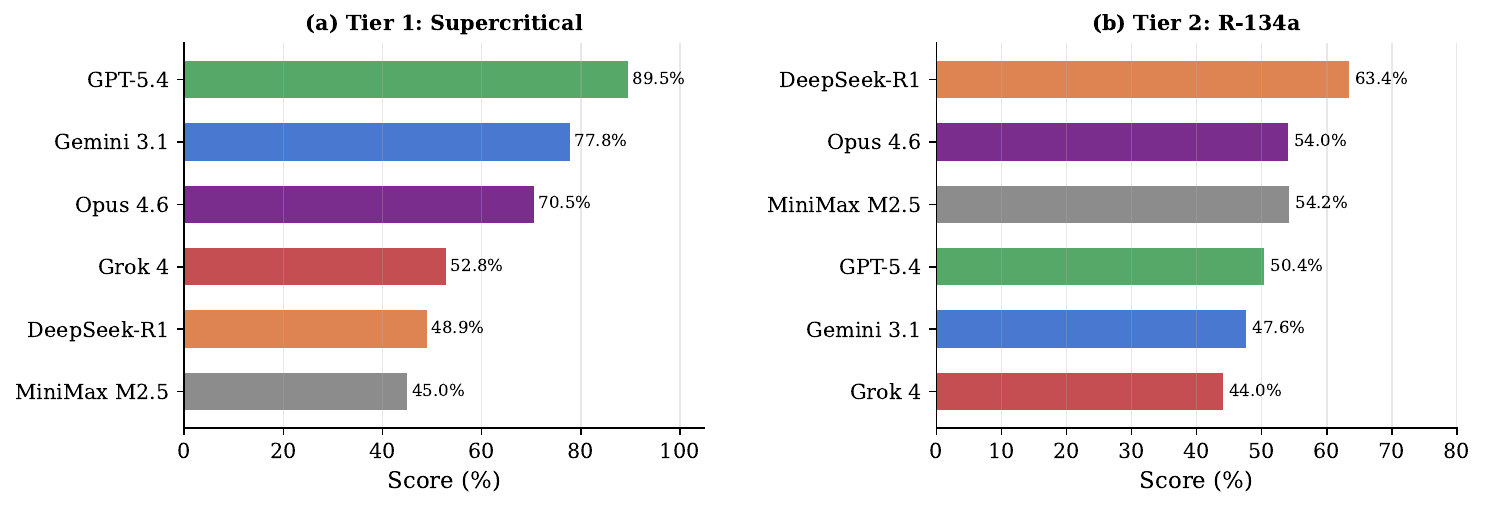}
\caption{Natural discriminators. (a)~Supercritical water properties in Tier~1: 45--90\% spread. (b)~R-134a refrigerant in Tier~2: 44--63\%. Faded bars show the gap to overall tier score.}
\label{fig:discriminators}
\end{figure}

We identify question subsets that produce the widest model separation (Figure~\ref{fig:discriminators}).

\textbf{Tier 1: Supercritical water.} The 10 supercritical questions ($T > 373.95^\circ$C, $P > 22.064$~MPa) produce a 44.5$\pp$ spread (GPT-5.4 at 89.5\% vs.\ MiniMax at 45.0\%). Models memorize steam table values from textbooks but cannot interpolate near the critical point where properties change extremely nonlinearly. Example: at 402$^\circ$C and 25.3~MPa, one model interpolated $h = 1{,}887$~kJ/kg from memorized values; IAPWS-IF97 gives $h = 2{,}585.8$~kJ/kg---a 27\% error.

\textbf{Tier 2: R-134a refrigerant.} All models collapse on R-134a (44--63\%), compared to 75--98\% on water. Training data is overwhelmingly steam tables; refrigerant property knowledge is scarce. The compressor component amplifies this: the work input formula $w_\text{in} = (h_{2s} - h_1) / \eta_s$ requires dividing by isentropic efficiency (not multiplying, as for turbines), and models frequently reverse this.

\textbf{Tier 3: Hard cycle variants.} BRY-AV (variable-$c_p$ aftercooling Brayton), BRY-RV (variable-$c_p$ regenerative Brayton), and CCGT (combined cycle) produce the widest T3 spreads. Opus scores 93.7\% mean on these hard variants; MiniMax scores 31.4\%.

\subsection{Component-Level Analysis}

\begin{figure}[!htb]
\centering
\includegraphics[width=0.95\linewidth]{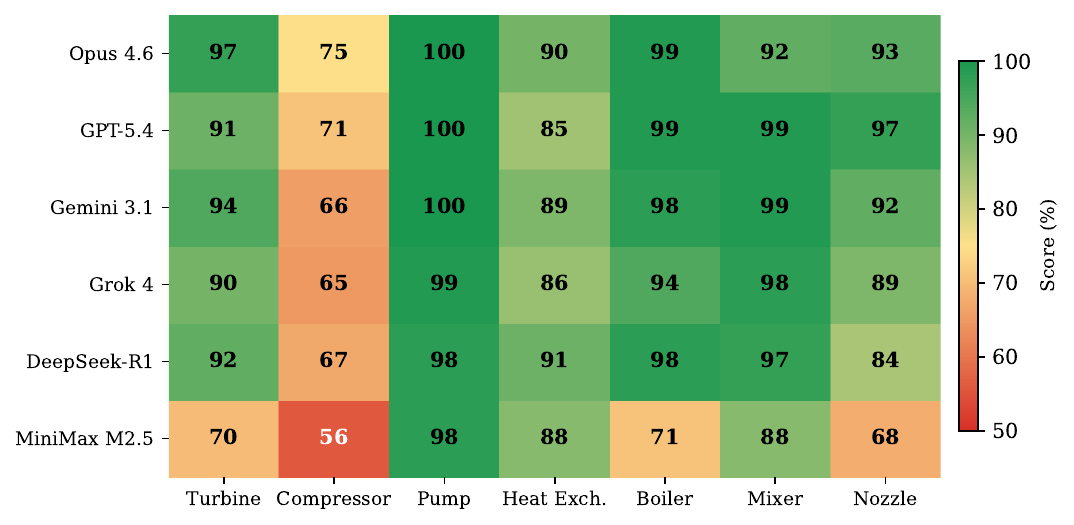}
\caption{Tier~2 component heatmap (3-run mean scores). The compressor column is uniformly the hardest; pump is near-trivial for all models.}
\label{fig:components}
\end{figure}

Figure~\ref{fig:components} reveals the compressor as the universally hardest component (55--75\%), while pumps are near-trivial ($\geq$97.5\%). Even the top-ranked model (Opus) scores only 75\% on compressors---far below its 92\% overall Tier~2 average. The difficulty arises from three compounding factors: (1)~the division formula for isentropic efficiency, (2)~double interpolation required to find $h_{2s}$ from $(s_1, P_2)$, and (3)~most R-134a questions target compressors, creating a dual penalty.

\FloatBarrier
\subsection{Cycle-Type Performance}

\begin{figure}[!htb]
\centering
\includegraphics[width=0.95\linewidth]{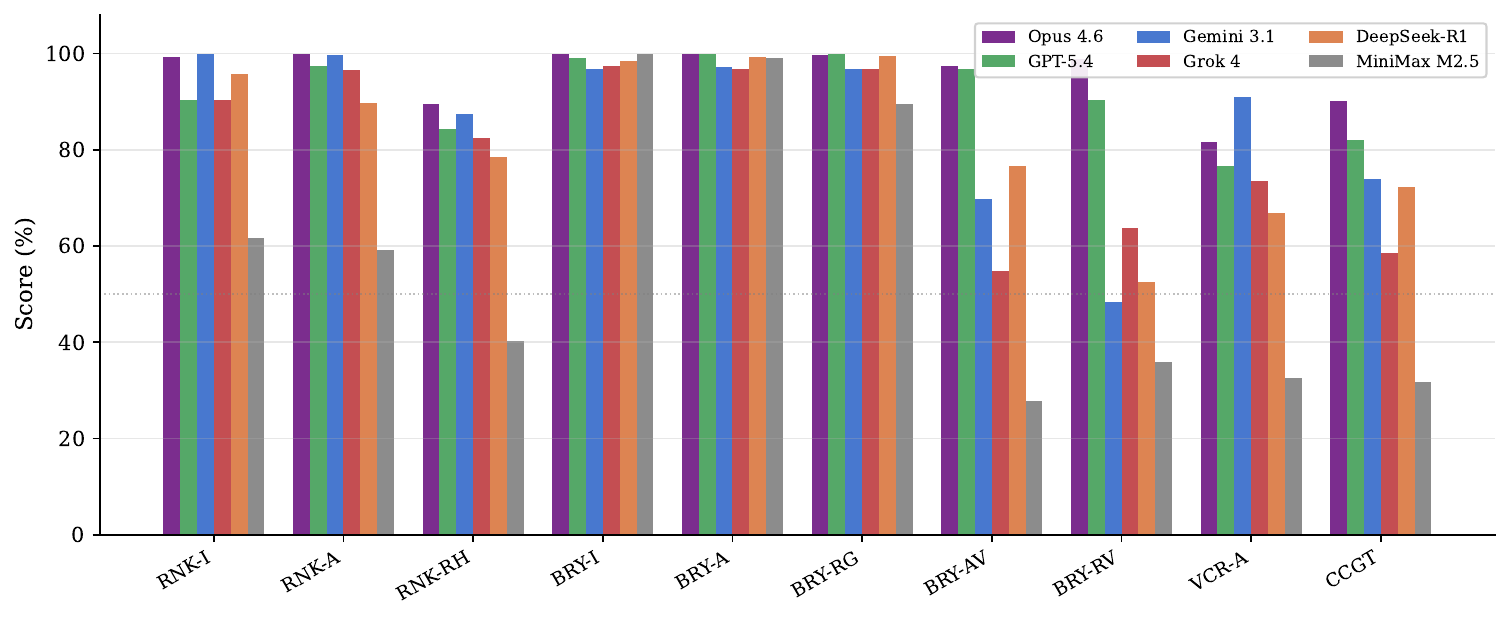}
\caption{Tier~3 scores by cycle type. Ideal Brayton variants (BRY-I, BRY-A, BRY-RG) are solved near-perfectly by frontier models. Variable-$c_p$ variants (BRY-AV, BRY-RV), VCR, and CCGT provide meaningful discrimination.}
\label{fig:cycles}
\end{figure}

Figure~\ref{fig:cycles} shows performance across all 10 cycle types. Ideal Brayton cycles (BRY-I/A/RG) are solved near-perfectly by all models except MiniMax, confirming that constant-$c_p$ ideal-gas calculations are well-represented in training data. The diagnostic value comes from:

\begin{itemize}[leftmargin=*,topsep=2pt,itemsep=1pt]
\item \textbf{BRY-AV/BRY-RV}: Variable-$c_p$ Brayton analysis requires the relative pressure ($P_r$) method from air tables, where isentropic relations take the form $P_2/P_1 = P_r(T_2)/P_r(T_1)$ rather than the constant-$c_p$ formula $T_2/T_1 = (P_2/P_1)^{(k-1)/k}$. The latter is invalid when $k$ varies with temperature. Gemini scores 69.7\% / 48.4\% on these variants---its responses consistently apply constant-$c_p$ isentropic formulas despite explicit ``variable specific heats'' instructions in the problem statement.
\item \textbf{CCGT}: Combined-cycle analysis couples a gas turbine (Brayton) with a steam turbine (Rankine) through a heat recovery steam generator (HRSG). With 9 state points across two coupled fluids, errors in the gas cycle propagate to the steam cycle via the HRSG energy balance. Best model: 90.1\% (Opus).
\item \textbf{VCR-A}: Vapor-compression refrigeration with R-134a exposes both refrigerant property gaps and the throttling process (isenthalpic expansion, $h_3 = h_4$, not isentropic), which models frequently mischaracterize.
\end{itemize}

\FloatBarrier
\subsection{Depth Progression}

\begin{figure}[!htb]
\centering
\includegraphics[width=0.95\linewidth]{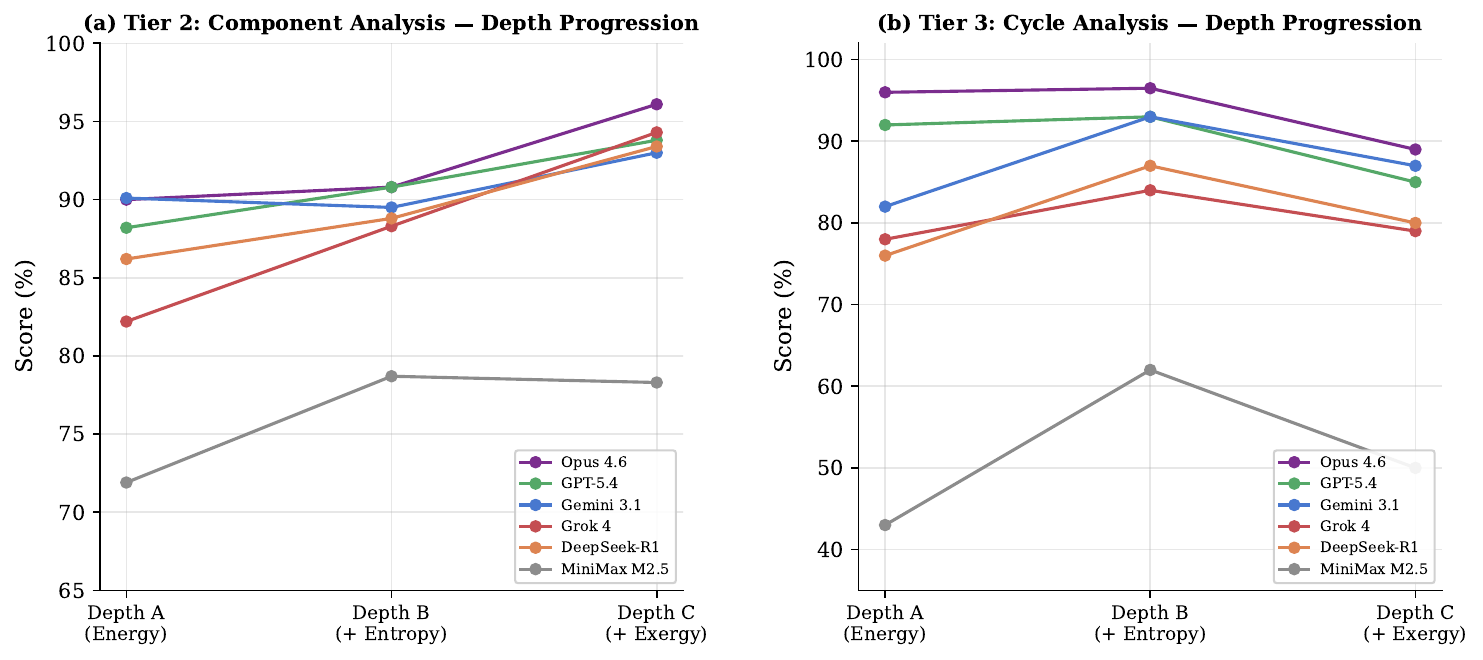}
\caption{Performance by analysis depth. (a)~Tier~2: counter-intuitive improvement at Depth~C for frontier models. (b)~Tier~3: peak at Depth~B, drop at Depth~C.}
\label{fig:depth}
\end{figure}

A surprising finding (Figure~\ref{fig:depth}a) is that frontier models score \emph{higher} on Depth~C (exergy analysis) than Depth~A (energy balance) in Tier~2. For example, Grok~4 scores 82.2\% on Depth~A but 94.3\% on Depth~C---a 12.1$\pp$ improvement as difficulty supposedly increases. We attribute this to the scoring design: Depth~C adds exergy steps with high weights, and these exergy calculations are often straightforward formulas ($\dot{X}_\text{dest} = T_0 \dot{S}_\text{gen}$, $\eta_\text{II} = 1 - \dot{X}_\text{dest}/\dot{X}_\text{in}$) applied to values the model has already computed. The additional steps provide more opportunities for partial credit. In Tier~3, the pattern reverses at Depth~C as the exergy calculations become more complex.

\FloatBarrier
\subsection{Multi-Run Consistency}

\begin{figure}[!htb]
\centering
\includegraphics[width=0.85\linewidth]{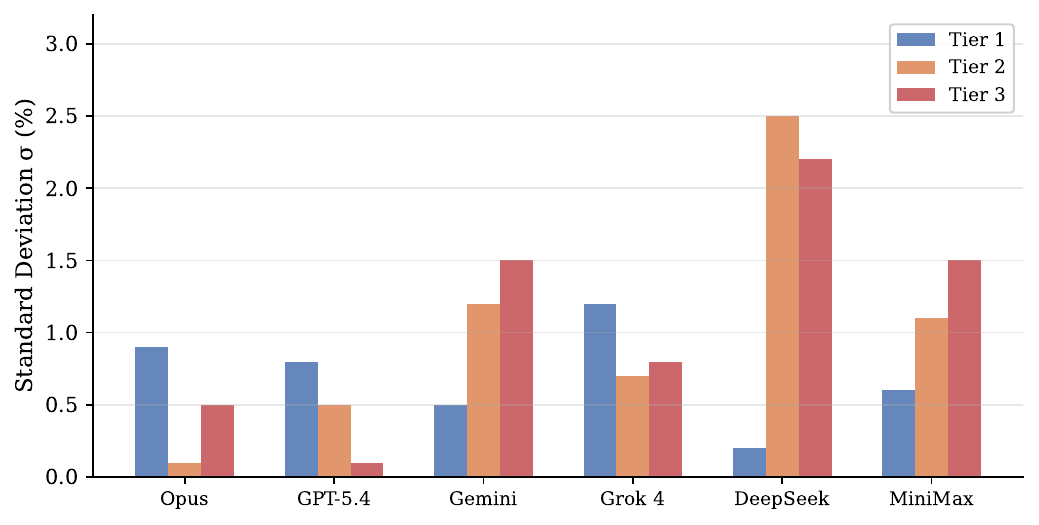}
\caption{Standard deviation across 3 independent runs. DeepSeek-R1 shows the highest variance on Tiers~2--3; GPT-5.4 is the most consistent on Tier~3.}
\label{fig:consistency}
\end{figure}

Figure~\ref{fig:consistency} reveals that run-to-run variability differs by an order of magnitude across models. DeepSeek-R1 achieves the lowest $\sigma$ on Tier~1 ($\pm$0.2\%) but the highest on Tier~2 ($\pm$2.5\%). GPT-5.4 is near-deterministic on Tier~3 ($\sigma = \pm$0.1\%). This consistency dimension is invisible to single-run evaluations and has practical implications: a model with $\pm$2.5\% variance is less reliable for engineering applications than one with $\pm$0.1\%, even if their mean scores are similar.

Pairwise Welch's $t$-tests on Tier~3 scores show that most model pairs are statistically distinguishable ($p < 0.05$) with only 3 runs. Notable exceptions: GPT-5.4 vs.\ Gemini ($t = 2.49$, not significant) and Grok~4 vs.\ DeepSeek ($t = -0.49$, not significant), indicating these pairs require more runs to separate reliably.

\FloatBarrier
\subsection{Token Efficiency}

\begin{figure}[!htb]
\centering
\includegraphics[width=0.75\linewidth]{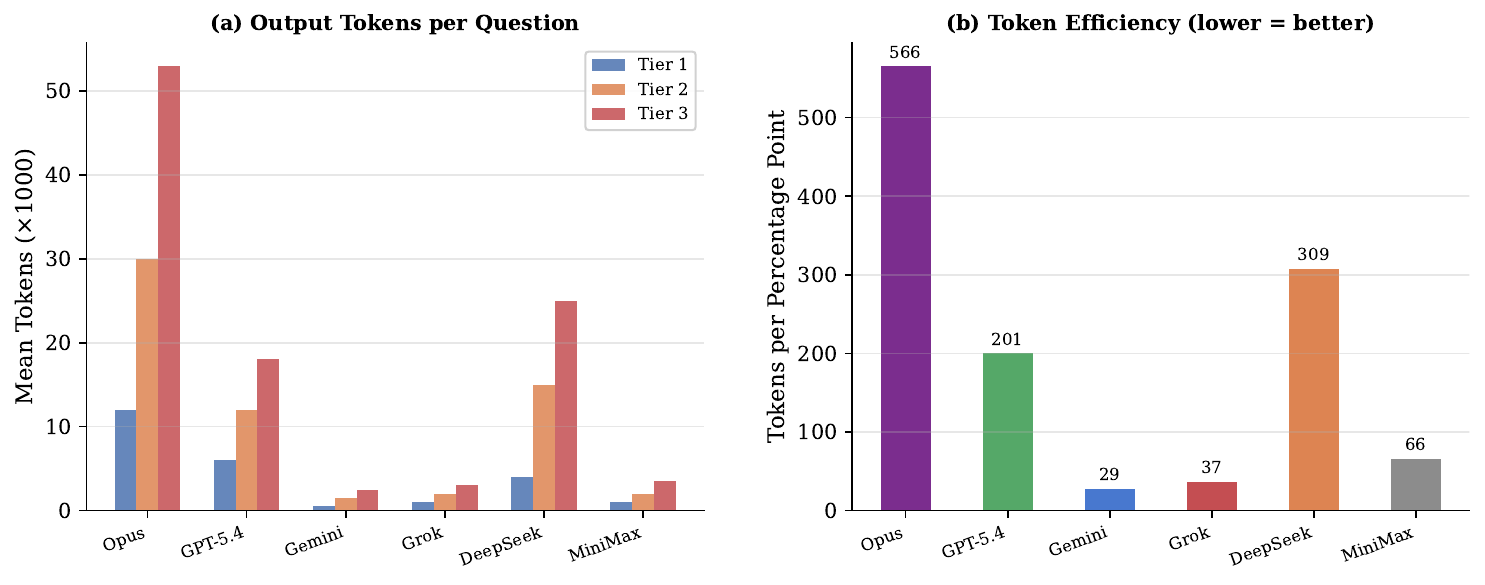}
\caption{Token usage and efficiency across all three tiers. (a)~Mean output tokens per question (including thinking tokens where exposed by the API). Opus invests up to 53K tokens on Tier~3; Gemini and Grok~4 stay below 3K. (b)~Tokens per percentage point on Tier~3 (lower = more efficient), showing a 20$\times$ gap between Opus and Gemini.}
\label{fig:tokens}
\end{figure}

Figure~\ref{fig:tokens} shows output token usage and efficiency across all three tiers. Opus invests heavily in reasoning---12K tokens on Tier~1 rising to 53K on Tier~3 (4.4$\times$ increase)---but achieves the smallest accuracy drop ($-$2.8$\pp$). Gemini and Grok~4 produce answers in $<$3K tokens, roughly a 20$\times$ cost advantage over Opus on Tier~3 for a 6$\pp$ accuracy trade-off. Token efficiency (tokens per percentage point) reveals diminishing returns: Opus requires $\sim$566~tok/pp on Tier~3, while Gemini achieves $\sim$29~tok/pp---a 20$\times$ gap for a modest accuracy difference. The Pearson correlation between tokens and accuracy increases from $r = 0.07$ (Tier~1) to $r = 0.29$ (Tier~3), suggesting that additional reasoning effort becomes more valuable as problem complexity grows, but the relationship remains weak. This aligns with the observation by \citet{loubet2025superstudent} that reasoning-enabled models (o3) show qualitative capability jumps on thermodynamic problems.

\subsection{Error Pattern Analysis}

Error propagation analysis reveals strong cascading: the $\phi$ coefficient between adjacent step pairs reaches 0.94 for $h_{2s} \to h_2$ (Tier~3), confirming that an error in the isentropic exit state propagates to the actual exit state. The dominant error mode across all models is ``missing'' (value not provided) rather than ``out of tolerance'' (wrong value)---models that attempt a calculation step almost always produce an answer within 2\% tolerance, but frequently omit steps entirely.

\FloatBarrier
\section{Discussion}
\label{sec:discussion}

\paragraph{The supercritical boundary: where memorization breaks down.} The most striking single finding in \ThermoQA{} is the supercritical water category, where all models collapse. The 10 supercritical questions produce a 44.5$\pp$ performance spread---the widest of any category in the benchmark---with even the best model (GPT-5.4) scoring only 89.5\% and the weakest at 45.0\%. The failure mechanism is clear: LLMs have memorized discrete steam table entries at textbook pressure intervals (typically 1, 5, 10, 15, 20~MPa) but have not internalized continuous equations of state. Near the critical point ($T_c = 373.95^\circ$C, $P_c = 22.064$~MPa), thermodynamic properties change extremely nonlinearly---small changes in temperature or pressure produce large changes in enthalpy, entropy, and density. Naive interpolation between memorized values that are far apart in property space produces catastrophic errors: the 27\% enthalpy error documented in Section~\ref{sec:results} ($h = 1{,}887$ vs.\ $2{,}585.8$~kJ/kg) is not a rounding issue but a fundamental failure of the interpolation strategy. This finding has broader implications for LLM evaluation in any quantitative domain: high accuracy on well-sampled regions of a function does not guarantee reliability in regions where the function is highly nonlinear, and benchmarks should deliberately probe these boundary zones.

\paragraph{Memorization vs.\ reasoning across tiers.} The ranking reshuffle from Tier~1 (Gemini \#1) to Tier~3 (Opus \#1) confirms that property retrieval and multi-step engineering reasoning are separable skills. Three recurring error patterns illustrate this disconnect. First, models reverse the compressor isentropic efficiency formula---applying $w_\text{in} = (h_{2s} - h_1) \times \eta_s$ instead of the correct $(h_{2s} - h_1) / \eta_s$---suggesting they have memorized a generic ``efficiency times ideal work'' template from turbine problems without understanding the physical asymmetry between work-producing and work-consuming devices. Second, models apply constant-$c_p$ isentropic relations ($T_2/T_1 = (P_2/P_1)^{(k-1)/k}$) even when the problem explicitly specifies variable specific heats and instructs use of air tables. This indicates rigid pattern-matching to the more frequently encountered constant-$c_p$ case rather than conditional selection of the appropriate method. Third, in multi-step cycle analysis, models occasionally produce thermodynamically inconsistent results---for example, reporting a turbine work output that does not match $h_1 - h_2$ given their own stated enthalpy values---revealing a failure to self-verify intermediate results against conservation laws.

These are not random errors; they are systematic reasoning failures that persist across runs, suggesting structural gaps in how models represent thermodynamic knowledge rather than stochastic sampling artifacts. The open-ended numerical format makes these gaps visible in ways that multiple-choice benchmarks cannot: a model must produce $h = 2585.77$~kJ/kg, not select it from four options.

\paragraph{Training data bias: water vs.\ refrigerants.} The fluid-specific results reveal a striking asymmetry in LLM training data. On Tier~2, all models score 75--98\% on water but collapse to 44--63\% on R-134a. Water/steam tables appear ubiquitously in textbooks, online resources, and solved examples; refrigerant property data is far less represented. This training data bias has practical implications: vapor-compression refrigeration is the most widely deployed thermodynamic cycle in the world (every air conditioner, refrigerator, and heat pump), yet the models are least equipped to analyze it. The IIR reference state convention for R-134a ($h = 200$~kJ/kg, $s = 1.0$~kJ/(kg$\cdot$K) at $0^\circ$C saturated liquid) adds a further challenge: models trained predominantly on ASHRAE or NBP conventions produce systematically offset property values even when the IIR convention is stated explicitly in the problem.

\paragraph{Error cascading in multi-step analysis.} The $\phi = 0.94$ error correlation between $h_{2s}$ and $h_2$ in Tier~3 reveals that thermodynamic calculations are uniquely sensitive to cascade failures. Unlike mathematics benchmarks where each step is relatively independent, in engineering thermodynamics the isentropic exit state anchors all subsequent calculations: actual exit enthalpy, work output, entropy generation, and exergy destruction all derive from it. A single property lookup error at state~2s can invalidate an entire cycle analysis. This explains why the dominant error mode is ``missing'' rather than ``wrong''---when models do attempt a calculation, they usually get within 2\% tolerance, but the chain breaks when they cannot determine an intermediate state and skip subsequent steps entirely.

\paragraph{Why we test without tools.} We deliberately evaluate without tool access because \ThermoQA{} measures what models \emph{know}, not what they can compute given the right tools. In preliminary experiments, enabling CoolProp via function calling raised Tier~1 scores to near 100\% for all models while leaving Tier~3 scores largely unchanged---confirming that Tier~1 errors are property retrieval failures and Tier~3 errors are reasoning failures. This observation validates our tier design: the tiers genuinely separate distinct capabilities. A systematic tool-augmented evaluation track is left to future work, but we note that tool-augmented models would see disproportionate gains on property-dependent steps, since the reasoning steps downstream are often executed correctly once anchor state values are right.

\paragraph{Consistency as an engineering requirement.} The order-of-magnitude range in run-to-run $\sigma$ across models ($\pm$0.1\% to $\pm$2.5\%) has direct implications for engineering deployment. A bridge designer using an LLM-assisted calculation tool needs the same answer every time, not an answer that varies by 2.5 percentage points between Tuesday and Wednesday. DeepSeek-R1 illustrates this tension: it achieves the most consistent Tier~1 scores ($\sigma = \pm$0.2\%) but the least consistent Tier~2 scores ($\sigma = \pm$2.5\%), suggesting that its reasoning chain is stable for simple lookups but introduces stochastic branching on multi-step problems. GPT-5.4, by contrast, maintains $\sigma = \pm$0.1\% on Tier~3---near-deterministic even on the most complex cycle calculations. We advocate that any numerical benchmark report $\sigma$ alongside mean accuracy; a model with 90\%~$\pm$~0.1\% is more useful in practice than one with 91\%~$\pm$~2.5\%.

\paragraph{The Depth~C scoring paradox.} The counter-intuitive finding that frontier models score higher on Depth~C (exergy) than Depth~A (energy) in Tier~2 warrants careful interpretation. We attribute this primarily to scoring mechanics rather than genuine exergy mastery: Depth~C questions include all Depth~A steps plus additional exergy steps ($\dot{X}_\text{dest} = T_0 \dot{S}_\text{gen}$, $\eta_\text{II} = 1 - \dot{X}_\text{dest}/\dot{X}_\text{in}$) that carry high weights and are straightforward formulas applied to previously computed values. This creates more opportunities for partial credit. The pattern reverses in Tier~3, where exergy calculations in full cycles are more complex and involve multiple components. Future benchmark iterations could address this by normalizing weights across depths or introducing depth-independent scoring.

\paragraph{Limitations.}
\begin{enumerate}[leftmargin=*,topsep=2pt,itemsep=1pt]
\item All problems are well-posed with unique numerical answers; real engineering involves design tradeoffs, incomplete specifications, and judgment calls that this benchmark does not capture.
\item No tool access is provided. While this isolates intrinsic model knowledge, it penalizes models that could perform well with equation-of-state solvers---a realistic engineering workflow.
\item Three runs provide limited statistical power; 95\% confidence interval widths range from 0.7 to 12.3\%. More runs would sharpen model separation, particularly for the GPT-5.4/Gemini and Grok~4/DeepSeek pairs that did not reach significance.
\item The $\pm$2\% scoring tolerance is standard for engineering coursework but may be too lenient for safety-critical applications such as nuclear or aerospace thermal design.
\item All questions are text-only. Diagram interpretation---reading T-s diagrams, P-h charts, or cycle schematics---is an essential engineering skill that this benchmark does not test.
\item While question parameters are generated programmatically (reducing memorization risk), the underlying problem structures follow standard textbook patterns; data contamination from training corpora cannot be fully excluded.
\item The benchmark was developed with assistance from Anthropic's Claude; however, all ground-truth values are computed independently via CoolProp, evaluation uses identical prompts across all providers, and scoring is fully deterministic. The first-place ranking of Claude Opus~4.6 should be interpreted with this disclosure in mind.
\end{enumerate}

\section{Conclusion}
\label{sec:conclusion}

\ThermoQA{} is a 293-question open-ended benchmark for engineering thermodynamics that tests property retrieval, component-level analysis, and full cycle calculations across four working fluids and three analysis depths. Evaluating six frontier LLMs across three independent runs each, we find that:

\begin{enumerate}[leftmargin=*,topsep=2pt,itemsep=2pt]
\item \textbf{Supercritical water exposes the limits of memorization.} The 10 supercritical questions produce the widest performance spread in the entire benchmark (44.5$\pp$). Models that score $>$97\% on standard steam table lookups drop to 45--90\% above the critical point, where properties change nonlinearly and memorized tabular values fail. This is the single most diagnostic category for distinguishing genuine thermodynamic knowledge from table memorization.

\item \textbf{Property lookup does not predict reasoning.} Rankings reshuffle from Tier~1 to Tier~3. The best memorizer (Gemini, 97.9\% on property lookups) is not the best reasoner (Opus, 93.6\% on cycle analysis). Cross-tier degradation ranges from 2.8$\pp$ to 32.5$\pp$, confirming these are separable skills.

\item \textbf{Real fluids and complex cycles expose training data gaps.} R-134a refrigerant (44--63\% across all models) and combined-cycle gas turbine analysis (31--90\% range) serve as natural discriminators, offering targeted diagnostic value for model developers.

\item \textbf{Reasoning consistency varies by an order of magnitude.} Multi-run $\sigma$ ranges from $\pm$0.1\% (GPT-5.4 on Tier~3) to $\pm$2.5\% (DeepSeek-R1 on Tier~2). Single-run evaluations miss this dimension entirely.

\item \textbf{Errors cascade through calculations.} A single property error at an isentropic exit state propagates through work, entropy, and exergy calculations with $\phi = 0.94$ correlation, making tool-augmented evaluation a natural next step.
\end{enumerate}

Several directions remain open. A tool-augmented track allowing CoolProp access would isolate reasoning errors from property retrieval failures and test models in a more realistic engineering workflow. Diagram interpretation---reading T-s and P-h charts---would add a multimodal dimension. The programmatic generation pipeline can produce new question instances indefinitely, providing resilience against data contamination. Evaluation of smaller and open-weight models (e.g., fine-tuned 8B parameter models for thermodynamic reasoning) would broaden the benchmark's utility beyond frontier systems. All data, code, and model responses are released at \url{https://github.com/olivenet-iot/ThermoQA}.

\section*{Acknowledgments}

This work was conducted independently. The author thanks the open-source CoolProp project \citep{coolprop} for validated thermodynamic property calculations. Development tooling included Anthropic's Claude for code assistance; all evaluation and scoring components are open-source and independently verifiable.

\FloatBarrier
\bibliographystyle{plainnat}

\appendix

\section{System Prompt}
\label{app:prompt}

All models receive the following identical system prompt. The question text is appended as the user message.

\begin{quote}
\small
\ttfamily
You are a thermodynamics expert. Answer the following thermodynamics question.
Show your reasoning, then report your final numerical answers clearly using standard symbols in exactly this format:

[symbol] = [value] [unit]

Examples:\\
h = 2943.12 kJ/kg\\
s = 6.9265 kJ/(kg\textperiodcentered K)\\
v = 0.1386 m\textthreesuperior/kg\\
x = 0.85\\
Phase: superheated vapor
\end{quote}

\section{Example Questions}
\label{app:examples}

\subsection*{Tier 1 Example (Supercritical)}

\begin{quote}
\small
\textit{At 402.0$^\circ$C and 25{,}300~kPa, find $h$, $s$, $v$, $u$, $\rho$, and the phase of water. Report your answers as: $h$ = \_\_\_ kJ/kg, $s$ = \_\_\_ kJ/(kg$\cdot$K), $v$ = \_\_\_ m$^3$/kg, $u$ = \_\_\_ kJ/kg, $\rho$ = \_\_\_ kg/m$^3$, Phase: \_\_\_}
\end{quote}

Ground truth (CoolProp, IAPWS-IF97): $h = 2585.77$~kJ/kg, $s = 5.1479$~kJ/(kg$\cdot$K), $v = 0.005994$~m$^3$/kg, $u = 2434.13$~kJ/kg, $\rho = 166.84$~kg/m$^3$. Phase: supercritical.

\subsection*{Tier 2 Example (Compressor, R-134a, Depth C)}

\begin{quote}
\small
\textit{R-134a enters a compressor at 200~kPa, $-10^\circ$C and exits at 1.2~MPa. The isentropic efficiency is 82\%. Using IIR reference state ($h = 200$~kJ/kg, $s = 1.0$~kJ/(kg$\cdot$K) at 0$^\circ$C saturated liquid), determine: (a)~$h_1$, $s_1$, (b)~$h_{2s}$, (c)~$h_2$, (d)~$w_\text{in}$, (e)~$s_2$, (f)~$\dot{S}_\text{gen}$, (g)~$\dot{X}_\text{dest}$, (h)~$\eta_\text{II}$. Dead state: $T_0 = 25^\circ$C, $P_0 = 100$~kPa.}
\end{quote}

This question has 8 scored steps with weights ranging from 1 (inlet properties) to 6 (second-law efficiency). All ground truth values are computed from CoolProp's Helmholtz EOS for R-134a.

\subsection*{Tier 3 Example (CCGT, Depth A)}

\begin{quote}
\small
\textit{A combined-cycle gas turbine plant operates with: compressor inlet $T_1 = 300$~K, $P_1 = 100$~kPa, compression ratio $r_p = 10$, combustion exit $T_3 = 1400$~K, gas turbine $\eta_s = 88\%$, compressor $\eta_s = 85\%$. The exhaust enters an HRSG producing steam at 6~MPa, 400$^\circ$C from feedwater at 30$^\circ$C, 6~MPa. Steam turbine $\eta_s = 90\%$, condenser at 10~kPa. Use variable specific heats (air tables) for the gas side.}
\end{quote}

\section{Weight Scheme}
\label{app:weights}

Table~\ref{tab:weights} shows the weight assignment for a Tier~3 CCGT question (Depth~A, 23 scored steps). State-point properties carry $w=1$; component-level calculations carry $w=3$; final cycle outcomes carry $w=5$.

\begin{table}[h]
\centering
\caption{Weight scheme for a CCGT Depth~A question (23 steps). The weighted score emphasizes engineering outcomes over intermediate state lookups.}
\label{tab:weights}
\small
\begin{tabular}{@{}llcl@{}}
\toprule
\textbf{Step} & \textbf{Description} & \textbf{Weight} & \textbf{Category} \\
\midrule
$h_1$, $h_{2s}$, $h_2$, $T_2$ & Gas compressor states & 1 & State points \\
$h_3$, $h_{4s}$, $h_4$, $T_4$ & Combustion \& turbine states & 1 & State points \\
$h_5$ & Gas turbine exhaust & 1 & State points \\
$h_6$, $h_{7s}$, $h_7$ & Steam cycle states & 1 & State points \\
$h_8$, $h_{9s}$, $h_9$ & Condenser/pump states & 1 & State points \\
\midrule
$\dot{m}_\text{steam}$ & HRSG mass balance & 3 & Component \\
$w_\text{comp}$, $q_\text{comb}$ & Gas-side work/heat & 3 & Component \\
$w_\text{gas\_turb}$ & Gas turbine work & 3 & Component \\
$w_\text{pump}$, $w_\text{steam\_turb}$ & Steam-side work & 3 & Component \\
\midrule
$\dot{W}_\text{net,combined}$ & Combined net power & 5 & Final outcome \\
$\eta_\text{combined}$ & Combined thermal efficiency & 5 & Final outcome \\
\bottomrule
\end{tabular}
\end{table}

\section{Evaluation Pipeline Detail}
\label{app:pipeline}

The extraction pipeline operates in two passes on each model response:

\paragraph{Pass 1: Regex extraction.} For each expected output variable, a set of regular expression patterns is applied to the response text. Patterns match standard engineering notation (e.g., \texttt{h = 2943.12 kJ/kg}), LaTeX subscripts (e.g., \texttt{h\_\{2s\}}), Unicode subscripts, and prose formats (e.g., ``the enthalpy is approximately 2943''). When multiple matches occur, the \emph{last} match in the text is retained, as LLMs typically state final answers after intermediate calculations. Auto-conversion heuristics handle common unit mismatches: values $>$10{,}000 for enthalpy are assumed to be in J/kg and divided by 1{,}000; efficiency values $>$1.0 are assumed to be percentages and divided by 100.

\paragraph{Pass 2: LLM extraction.} The full response text (including any thinking/reasoning tokens exposed by the API) is sent to gpt-4.1-mini at temperature=0 with a structured prompt requesting JSON output of the expected variable names and values. The extractor is instructed to return only final answers, not intermediate calculations.

\paragraph{Take-max combination.} For each scored step, the pipeline retains whichever extraction (regex or LLM) yields a correct answer when compared against ground truth. If both are correct, the LLM extraction is preferred. If neither is correct, the LLM extraction is used. This ensures the combined pipeline never scores lower than regex alone. In practice, LLM extraction improved aggregate scores by 1--3\% over regex alone, with take-max preventing the $\sim$0.5\% regression cases where the LLM extractor misinterpreted intermediate values as final answers.

\section{Statistical Significance}
\label{app:stats}

Pairwise Welch's $t$-test results on Tier~3 scores (3 runs per model):

\begin{table}[h]
\centering
\small
\begin{tabular}{@{}lcccccc@{}}
\toprule
& Opus & GPT-5.4 & Gemini & Grok 4 & DeepSeek & MiniMax \\
\midrule
Opus & --- & \checkmark & \checkmark & \checkmark & \checkmark & \checkmark \\
GPT-5.4 & & --- & $\times$ & \checkmark & \checkmark & \checkmark \\
Gemini & & & --- & \checkmark & \checkmark & \checkmark \\
Grok 4 & & & & --- & $\times$ & \checkmark \\
DeepSeek & & & & & --- & \checkmark \\
\bottomrule
\end{tabular}

{\footnotesize \checkmark = significant at $p < 0.05$; $\times$ = not significant.}
\end{table}

\end{document}